\documentclass[letterpaper]{article} 
\usepackage{aaai2026}  
\usepackage{times}  
\usepackage{helvet}  
\usepackage{courier}  
\usepackage[hyphens]{url}  
\usepackage{graphicx} 
\usepackage{natbib}  
\usepackage{caption} 
\usepackage{algorithm}
\usepackage{algorithmic}

\usepackage{booktabs}
\usepackage{multirow}
\usepackage{graphicx}
\usepackage{array}
\usepackage{amsmath}

\usepackage{newfloat}
\usepackage{listings}
\DeclareCaptionStyle{ruled}{labelfont=normalfont,labelsep=colon,strut=off} 
\floatstyle{ruled}
\newfloat{listing}{tb}{lst}{}
\floatname{listing}{Listing}
\title{Cross-Temporal 3D Gaussian Splatting for Sparse-View Guided Scene Update}
\author{
    Zeyuan An\textsuperscript{\rm 1},
    Yanghang Xiao\textsuperscript{\rm 1},
    Zhiying Leng\textsuperscript{\rm 1},
    Frederick W. B. Li\textsuperscript{\rm 2},
    Xiaohui Liang\textsuperscript{\rm 1,\rm 3}\thanks{Corresponding author.}
}

\affiliations{
    \textsuperscript{\rm 1}State Key Laboratory of Virtual Reality Technology and Systems, Beihang University, Beijing, China\\
    \textsuperscript{\rm 2}Department of Computer Science, University of Durham, U.K.\\
    \textsuperscript{\rm 3}Zhongguancun Laboratory, Beijing, China\\
    \{947066339, xiaoyanghang, zhiyingleng, liang\_xiaohui\}@buaa.edu.cn,
    frederick.li@durham.ac.uk
}

\usepackage{bibentry}

\begin{document}

\maketitle

\begin{abstract}
Maintaining consistent 3D scene representations over time is a significant challenge in computer vision. 
Updating 3D scenes from sparse-view observations is crucial for various real-world applications, including urban planning, disaster assessment, and historical site preservation, where dense scans are often unavailable or impractical. In this paper, we propose Cross-Temporal 3D Gaussian Splatting (Cross-Temporal 3DGS), a novel framework for efficiently reconstructing and updating 3D scenes across different time periods, using sparse images and previously captured scene priors. 
Our approach comprises three stages: 1) Cross-temporal camera alignment for estimating and aligning camera poses across different timestamps; 2) Interference-based confidence initialization to identify unchanged regions between timestamps, thereby guiding updates; and 3) Progressive cross-temporal optimization, which iteratively integrates historical prior information into the 3D scene to enhance reconstruction quality.
Our method supports non-continuous capture, enabling not only updates using new sparse views to refine existing scenes, but also recovering past scenes from limited data with the help of current captures. Furthermore, we demonstrate the potential of this approach to achieve temporal changes using only sparse images, which can later be reconstructed into detailed 3D representations as needed. Experimental results show significant improvements over baseline methods in reconstruction quality and data efficiency, making this approach a promising solution for scene versioning, cross-temporal digital twins, and long-term spatial documentation.
\end{abstract}


\section{Introduction}

\begin{figure}[t]
    \centering
    \includegraphics[width=1.\linewidth]{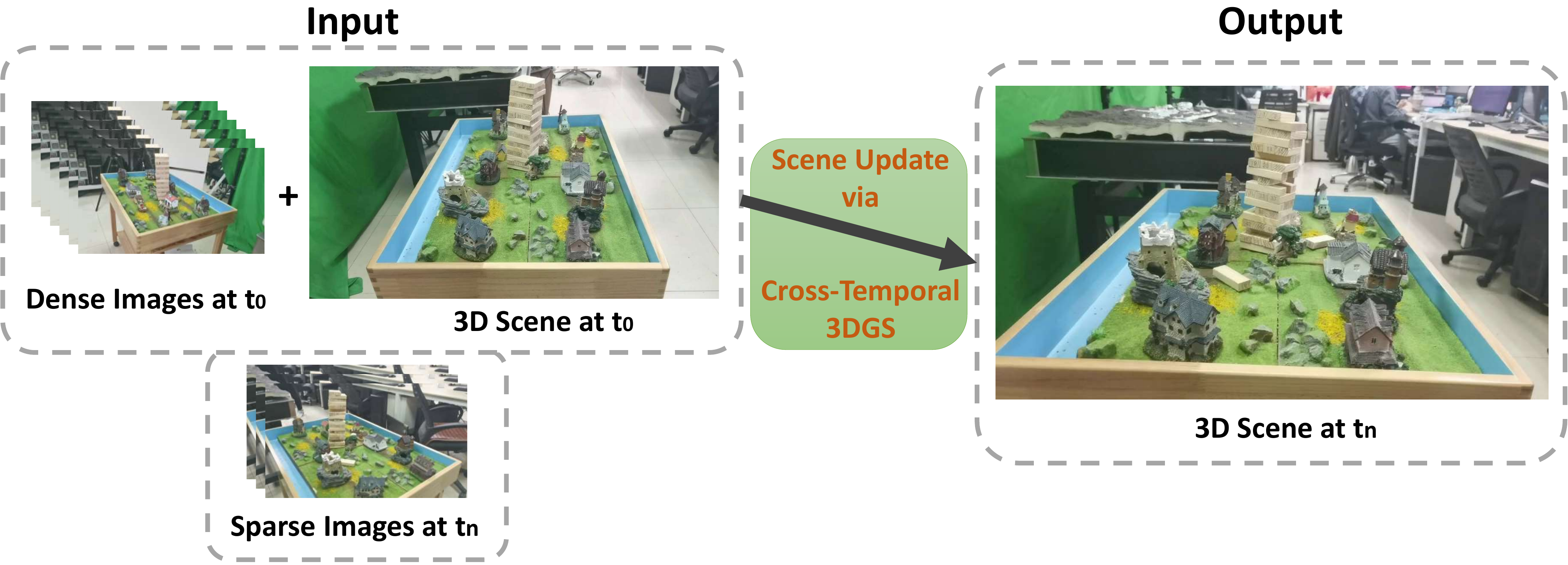}\\
    \caption{Scene update via Cross-temporal 3DGS. Given sparse-view inputs at $t_n$, our proposed Cross-temporal 3DGS updates the 3D scene from a well-observed timestamp $t_0$ to the sparse observed timestamp $t_n$.}
    \label{fig:main}
\end{figure}

Maintaining consistent and updatable 3D scene representations across time is a fundamental yet underexplored challenge in computer vision and graphics. While modern techniques such as Neural Radiance Fields (NeRF) and 3D Gaussian Splatting (3DGS) have dramatically improved the quality and interactivity of static 3D scene modeling, they are largely limited to single-timestamp reconstruction. In many real-world applications, such as urban digital twins, cultural heritage preservation and post-disaster response, there is a growing need to reconstruct, revise and compare scenes across different points in time, often from sparse or incomplete visual data. This demands not only high-fidelity scene reconstruction, but also robust temporal generalization under sparse-view and non-continuous capture settings.

A city planner may wish to assess changes in urban layout over several years using only archival images. A conservation team may need to restore a damaged historical site using both prior 3D reconstructions and newly acquired photographs. In these scenarios, data are often limited to a few unordered captures across different devices, viewpoints, or timestamps, making conventional dense-reconstruction pipelines infeasible. The ability to update scenes from sparse observations, using previously captured models as priors, would enable a “reconstruct-on-demand” paradigm for long-term spatial documentation and analysis.

Despite increasing interest in dynamic 3D reconstruction and scene editing, existing methods struggle to meet this goal. Dynamic 3DGS methods like 4D-GS require dense temporal observations and explicit motion modeling, rendering them unsuitable for independently captured timepoints. Change detection techniques such as Wild-Gaussian and CL-Splats can localize scene differences but often fail to reconstruct updated scene-level geometry or rely on camera consistency. Meanwhile, 3D scene editors such as GaussianEditor support manual or text-guided manipulation but depend heavily on user intervention and offer limited structural accuracy. These approaches expose a critical gap: the lack of a unified framework capable of updating 3D scenes from sparse cross-temporal images.

To address these challenges, we propose \textbf{Cross-Temporal 3D Gaussian Splatting (Cross-Temporal 3DGS)}, a novel framework designed to update or recover 3D scenes across time using sparse views and historical priors. Our method introduces a three-stage pipeline to bridge the temporal gap between independently captured observations:

Firstly, we implement \textbf{Cross-Temporal Camera Alignment}, which aligns camera poses by matching point clouds from \( t_0 \) and \( t_n \). This ensures geometric consistency across different capture conditions, laying a solid foundation for subsequent updates. 
The second stage is \textbf{Interference-Based Confidence Initialization}. Rather than depending on dense optical flow, we propose a mechanism that identifies regions in \( t_0 \) closely matching structures in \( t_n \). This approach generates confidence maps for the views in \( t_0 \), with high-confidence regions providing auxiliary supervision during reconstruction, thereby enhancing the reliability of updates.
Finally, we introduce \textbf{Progressive Cross-Temporal Optimization}. Utilizing the high-confidence regions from \( t_0 \), we conduct a confidence-guided 3DGS training at \( t_n \), where confidence maps are iteratively refined. This progressive integration of prior information helps mitigate the limitations posed by sparse observations at \( t_n \), ultimately improving both fidelity and efficiency.

We validate our approach through extensive experiments on both synthetic and real-world cross-temporal datasets. Compared to baseline, sparse-view reconstruction and editing methods, Cross-Temporal 3DGS achieves substantial improvements in reconstruction quality, structural consistency, and computational efficiency, demonstrating its potential for scalable deployment in temporal scene analysis and historical modeling tasks.

Our contributions can be summarized as follows:

\begin{itemize}
\item We introduce the first 3DGS-based framework capable of cross-temporal scene updates from sparse observations;
\item  We design a novel confidence-guided optimization strategy to selectively propagate reliable priors while adapting to new scene content;
\item We construct a benchmark dataset and show significant performance gains over baselines in quality, robustness, and efficiency.
\end{itemize}

\section{Related Works}

\textbf{}

\textbf{3D Gaussian Splatting:}
3D Gaussian Splatting (3DGS) has emerged as a powerful representation for high-quality 3D reconstruction and real-time rendering, offering superior performance compared to mesh-based methods and NeRF. Previous works has optimized 3DGS for efficiency \cite{navaneet2023compact3d,niedermayr2024compressed}, rendering fidelity \cite{cheng2024gaussianpro,yan2024multi,zhang2024fregs}, and sparse-view reconstruction \cite{chung2024depth,zhu2023fsgs}. Some works focus on real-time rendering \cite{fan2025fov, li2025mpgs}, while others target human reconstruction \cite{dongye2024adaptive, zhao2024gaussianhand, kleinbeck2025multi}. Also, several studies concentrate on scene reconstruction \cite{schieber2025semantics, wang2025look, zhai2025splatloc} and editing \cite{ren2024palettegaussian, schieber2025semantics}.

Compared to NeRF, which suffers from long training time and implicit representation, 3DGS provides explicit, splat-based structures that are more suitable for incremental updates, partial reconstructions, and scene comparison across time, which are the critical needs in urban planning, disaster inspection, and historical preservation. But most works assume single-timestamp or dense multi-view inputs, leaving a gap in addressing cross-temporal scene reconstruction from sparse observations.

Our work addresses this gap by leveraging 3DGS for non-continuous, temporally sparse scene updating, enabling historical priors to guide current reconstructions and supporting scene recovery even when only partial or intermittent imagery is available.

\textbf{3D Scene Updating:}
Despite recent advances in 3DGS, updating 3D scenes over time remains challenging, especially under sparse and temporally unaligned observations. Existing efforts can be categorized into three directions:

Dynamic reconstruction methods, such as 4D-GS \cite{wu20244d}, model continuous motion via per-frame deformation fields. Other studies focus on optimizing deformation fields for enhanced results \cite{lin2024gaussian, yang2024deformable}, while others model dynamic properties by constructing individual trajectories for each Gaussian \cite{kratimenos2024dynmf, li2024spacetime, luiten2024dynamic}. However, they rely on temporally dense sequences and accurate motion estimation, which are impractical for real-world scenarios with independently captured timepoints.

Change detection approaches, including 3DGS-CD \cite{lu20253dgs}, CL-Splats \cite{ackermann2025cl} and Fast-Nerf-Update \cite{lu2024fast}, focus on identifying local object-level changes. These methods are limited to foreground displacement or appearance variations and lack the capacity to recover complete scene structures across time. Other scene-level methods, such as WildGaussians \cite{kulhanek2024wildgaussians}, focus on the construction of stable segments of the scene, unable to perform scene updates.
Scene versioning and archival tools, like GS-LTS \cite{fu2025gs} and RealityGit \cite{li2023realitygit}, maintain temporal scene snapshots. However, they require strict camera pose consistency and often fail under sparse observations.
Some studies exploring semantic-driven editing \cite{haque2023instructnerf2nerf, palandra2024gsedit, brooks2023instructpix2pix,wang2023fg,wang2025fg,wang2025most,wang2024gaussianeditor, ren2024palettegaussian, schieber2025semantics}, supporting manual or text-driven edits but lack automatic change inference and often result in geometrically inconsistent updates. Others have utilized manually annotated objects for modifications \cite{gordon2023blended, zhuang2023dreameditor, chen2024gaussianeditor}, which also face the same problem. Our work benefits from these valuable works, while general deep learning training strategies \cite{bao2022minority,bao2022rethinking,bao2024improved,bao2025aucpro} also support our framework.

These limitations highlight the need for a framework that can (1) align independently captured scenes across time, (2) identify reliable priors without manual annotation, and (3) progressively update the scene from sparse images. We propose Cross-Temporal 3DGS, a unified pipeline for scene updates under sparse, asynchronous observations.


\begin{figure*}[t]
 \centering 
 \includegraphics[width=1.\linewidth]{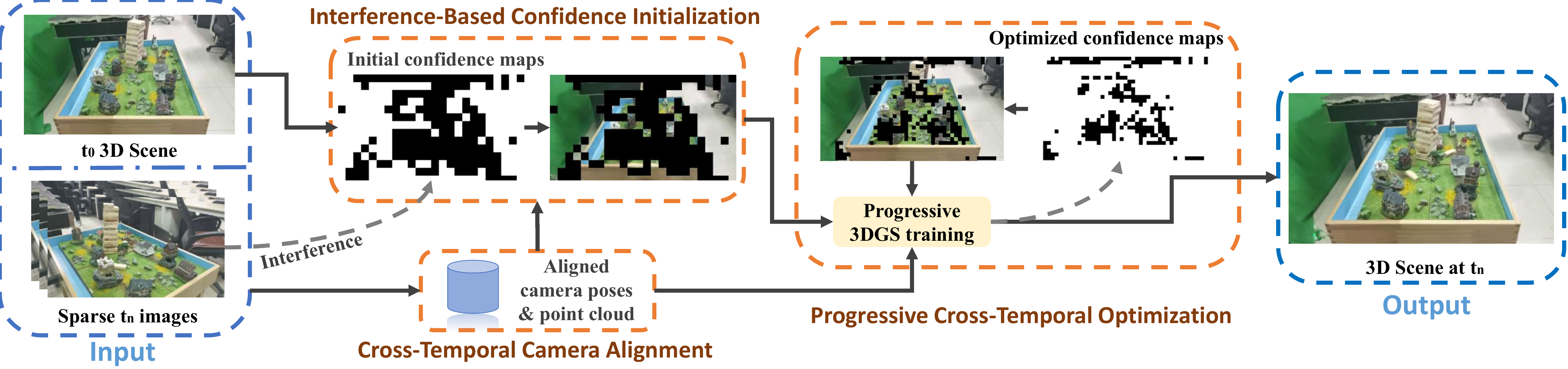}
 \caption{Overall pipeline of Cross-Temporal 3DGS.}
 \label{fig:overview}
\end{figure*}

\section{Methodology}

\textbf{Overview}
Directly reconstructing a 3D scene from sparse views at the target time results in significant degradation of reconstruction quality and fails to utilize historical data as effective priors, degrading the accuracy and usability of scene updates in real-world applications such as urban change monitoring or disaster recovery. Existing 3DGS-based scene update methods, such as scene editing, often rely on manual annotations or semantic guidance, which can be unreliable and uncontrollable. This limitation restricts usability and realism in real-world applications. Furthermore, current dynamic scene reconstruction methods based on 3DGS typically assume temporal continuity and camera consistency, necessitating extensive training time, making them ill-suited for scenarios where past and current data are captured under inconsistent conditions.

These limitations highlight several critical challenges for scene updates using 3DGS:
1) Lack of cross-temporal alignment for non-continuous captures, 
2) Inability to automatically extract valuable priors from previous observations, and
3) Failure to effectively leverage these priors during sparse-view reconstruction.

As shown in Figure \ref{fig:overview}, Our method addresses these gaps through three key innovations:
1) \textbf{Cross-Temporal Camera Alignment}: Utilizing point cloud registration to ensure robust alignment of camera poses across different timestamps, 
2) \textbf{Interference-Based Confidence Initialization}: Automatically identifying stable regions in the scene through interference-based methods, allowing for the preservation of prior knowledge, and
3) \textbf{Progressive Cross-Temporal Optimization}: Integrating high-quality scene priors into the 3DGS reconstruction process progressively, enhancing both fidelity and computational efficiency.

\subsection{Cross-Temporal Camera Alignment}
Independent pose estimation at separate time points can lead to coordinate system misalignment. 
To ensure geometric consistency across independently captured timestamps, we propose a cross-temporal camera alignment strategy that reconstructs and registers scene structures across time, enabling a unified coordinate for subsequent optimization.

The camera poses at different timestamps ($t_0$ and $t_n$) lead to an unknown rigid transformation (rotation $\mathbf{R}$ and translation $\mathbf{t}$) between their coordinate systems. Directly transferring prior scene information from $t_0$ to $t_n$ without addressing this misalignment introduces geometric inconsistencies and projection errors, ultimately degrading reconstruction quality. 
To resolve this issue, we align the two timestamps into a unified coordinate system by leveraging scene structures. While local scene variations may exist, we assume that the global layout remains largely unchanged, allowing us to determine an optimal rigid transformation using cross-temporal point cloud registration.

\begin{figure}[t]
 \centering 
 \includegraphics[width=0.7\columnwidth]{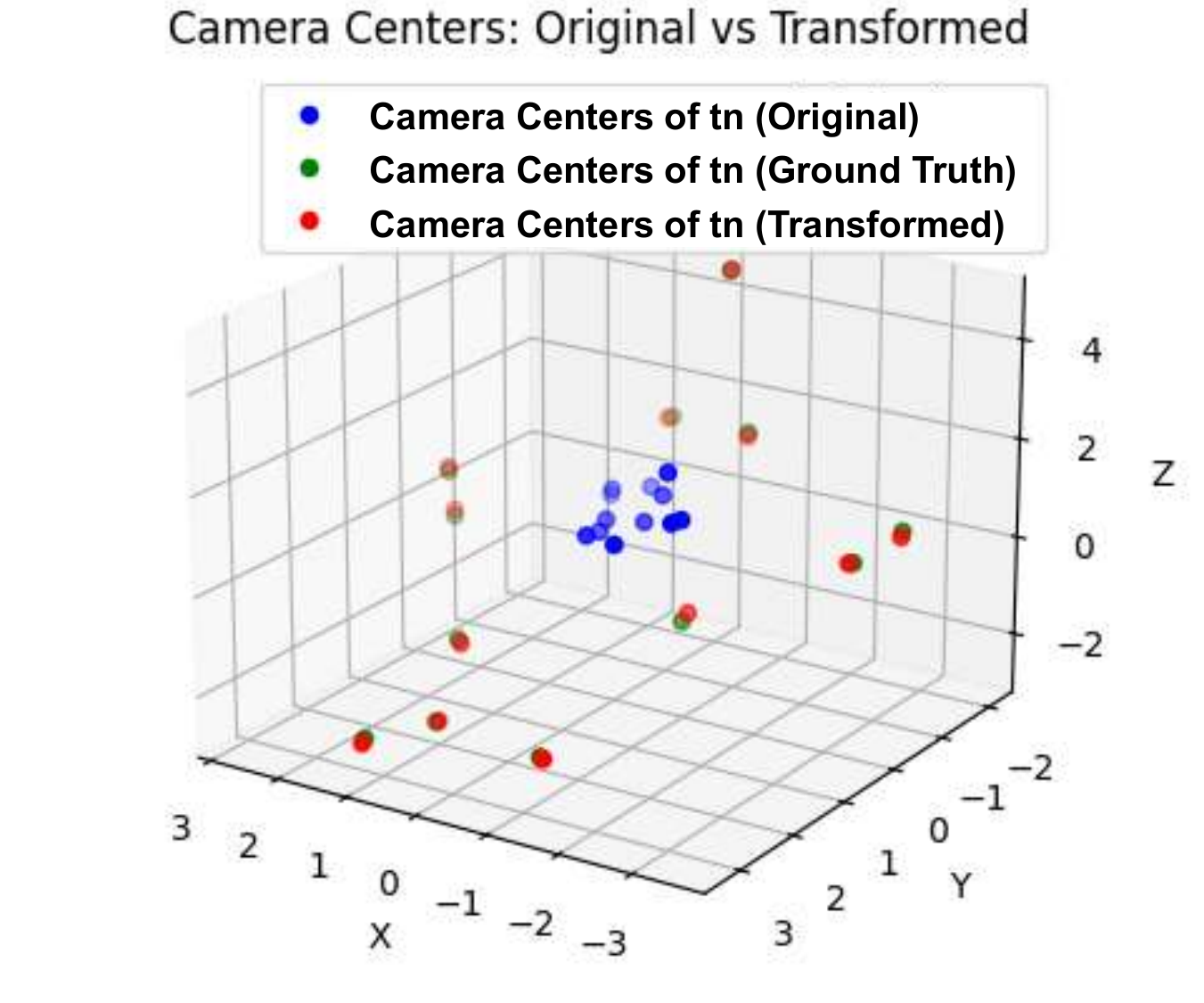}
 \caption{Cross-Temporal Camera Alignment. This process involves estimating and aligning camera poses from different timestamps (\(t_0\) and \(t_n\)) by registering dense point clouds, ensuring geometric consistency for effective scene updates.}
 \label{pic:camera}
\end{figure}

We estimate camera poses and generate a dense point cloud $P_n$ at $t_n$ using DUSt3R \cite{wang2024dust3r}, as dense point clouds show positive effects for sparse-view reconstruction \cite{fan2024instantsplat,chen2024dense}. Then downsample $P_n$ into 1/4 for better efficiency \cite{fan2024instantsplat}. To align these with the COLMAP coordinate system, we compute a similarity transformation $S$ by minimizing the reprojection error. This optimization is solved using the Levenberg–Marquardt (LM) algorithm \cite{triggs2000bundle}, which optimizes rigid transform parameters with 2D keypoint matches of points projected into $t_0$ views, where the reprojection error is the sum of squared pixel distances between DUSt3R point projections and 2D keypoint observations, minimized via LM to estimate the rigid transformation.

\begin{figure}[t]
 \centering 
 \includegraphics[width=0.9\columnwidth]{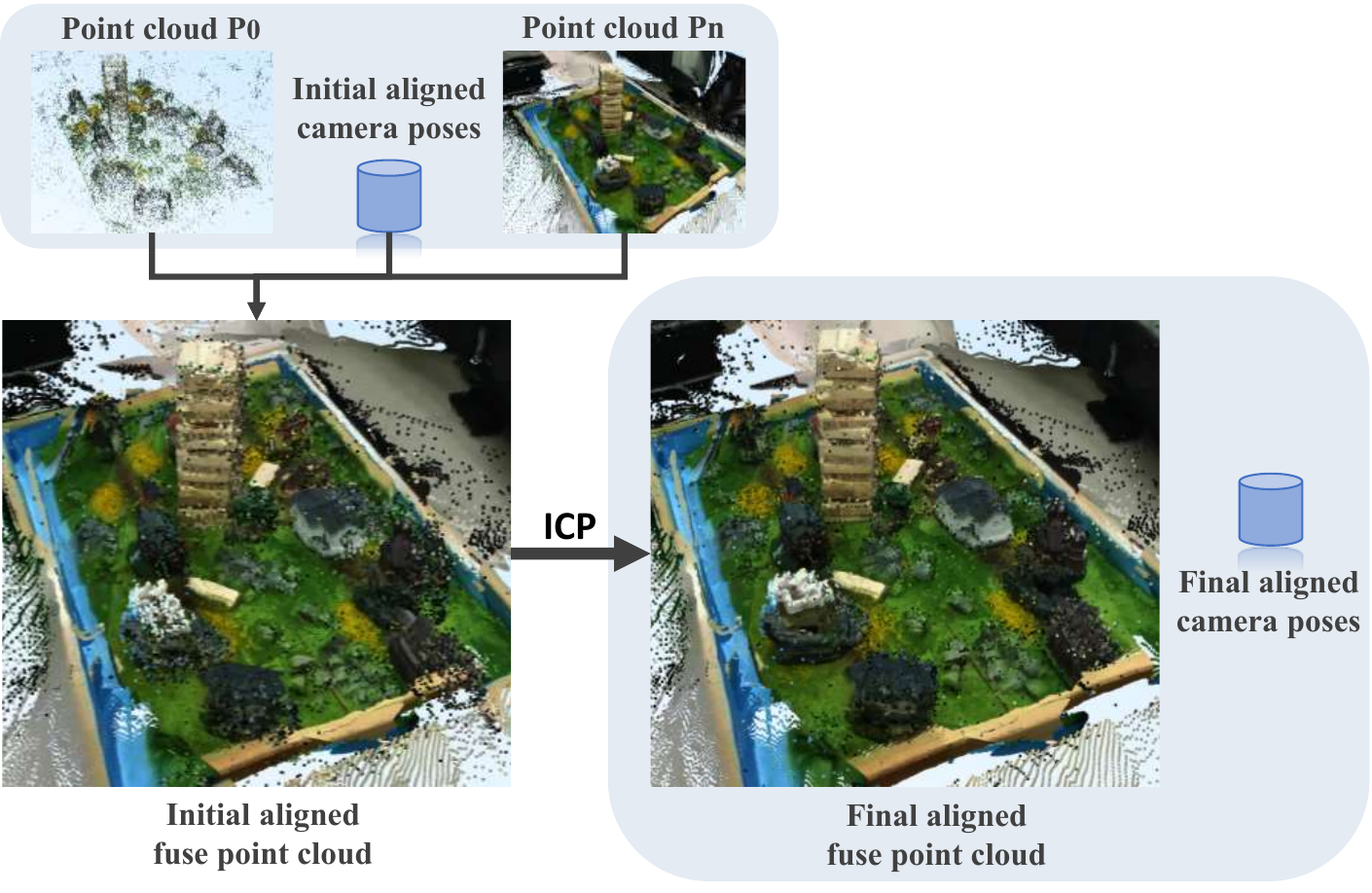}
 \caption{Point cloud alignment of $t_0$ and $t_n$. Initial alignment results (left) are further refined using ICP to achieve a more precise transformation (right), ensuring accurate registration of scene structures and camera poses.}
 \label{fig:method1}
\end{figure}

After obtaining the optimal transformation $S$, the Dust3R point cloud is transformed into an aligned point cloud as shown in the left-down of Figure \ref{fig:method1}. To further improve the geometric consistency between $P_n^{\text{aligned}}$ and the COLMAP point cloud $P_c$, we employ the Iterative Closest Point (ICP) algorithm \cite{zhang1994iterative} to formulate the transformation and apply to both the point cloud and the camera poses, the example result is in Figure \ref{pic:camera}.

Finally, the fused point cloud is defined as:
\begin{equation}
P_{\text{fused}} = P_c \cup P_n^{\text{aligned}}.
\label{eq:aa}
\end{equation}
This achieves accurate static regions while notably capturing the structural updates at $t_n$. This provides a robust geometric foundation for subsequent cross-temporal scene updates.

\subsection{Interference-Based Confidence Initialization}
Directly transferring the original 3DGS model can introduce artifacts in dynamic regions. To mitigate this, we propose a self-supervised confidence estimation method that automatically identifies stable regions, thereby reducing reliance on manual annotations. 

Our core insight is that relatively stable regions should demonstrate minimal degradation in rendering quality when their Gaussian parameters are subjected to controlled perturbations guided by views at $t_n$. Conversely, regions that experience significant changes between timestamps \(t_0\) and \(t_n\) will exhibit greater sensitivity to these perturbations. Based on this principle, we estimate confidence values for each region and construct a confidence map. This interference quantifies how the original scene remains stable.

We initiate the process using the 3D Gaussian model \(G_0\) reconstructed from \(t_0\). 
A brief adaptation phase is conducted with sparse-view images from \(t_n\), enabling the optimizer to adjust Gaussian parameters that conflict with the new observations, while allowing static regions to remain largely unchanged.

To quantify regional stability, we employ a modified Structural Similarity Index (SSIM) \cite{wang2004ssim} that minimizes the effects of luminance and contrast variations. We weight the perturbing $G_0$ by modified SSIM instead of direct pixel losses as the later mis-update unreliable regions (color similar or visible at $t_0$ but absent at $t_n$), causing artifacts. We calculate the modified SSIM between the rendered image \(I_r\) after perturbation and the original ground-truth image \(I_t\).
By excluding the luminance term and constraining the contrast term, this formulation emphasizes structural similarity, thereby reducing sensitivity to changes in illumination and contrast.

We construct confidence maps by partitioning the image space and computing a modified SSIM-based stability score. Regions exceeding a similarity threshold are treated as high-confidence areas and used to guide the update process.

\subsection{Progressive Cross-Temporal Optimization}
Under sparse-view conditions, 3DGS often lacks sufficient constraints, leading to floating artifacts and needle-like structures. To address this issue, we propose a progressive optimization strategy that utilizes the fused point clouds as geometric priors. Our method incrementally activates high-confidence regions first, followed by iterative confidence expansion, which allows low-confidence regions to be optimized in later stages. This approach ensures a more stable reconstruction by progressively integrating geometric priors.

We begin by using the fused point cloud \(P_{\text{fused}}\) to provide robust geometric constraints for the \(t_n\) scene. This point cloud leverages the high accuracy of the reconstruction from \(t_0\) alongside the updated structure from \(t_n\). The incorporation of such dense priors enhances 3DGS quality under sparse supervision and reduces the number of training iterations required. The fused point cloud serves solely as an initialization prior, gradually fading out as optimization progresses and image supervision dominates.

\subsubsection{Preliminary Training with Sparse Observations.}
Using \(P_{\text{fused}}\), we initialize the 3DGS model \(G_n\) at \(t_n\) and conduct a preliminary optimization utilizing sparse-view images from \(t_n\) and high-confidence regions from \(t_0\). The loss function combines photometric consistency with confidence-weighted regularization:
\begin{equation}
\begin{aligned}
L_{\text{init}} = \sum_{p} c_p \Bigg[
    &\sum_{k \in p} \left\| I_k^{\text{render}} - I_k^{t_n} \right\|_1  \\
    &+ \left( 1 - \text{SSIM}(I_p^{\text{render}}, I_p^{t_n}) \right)
\Bigg],
\end{aligned}
\end{equation}
where \(p\) denotes an image patch, \(c_p\) is the confidence value for patch \(p\) (ranging from 0 to 1), and \(k \in p\) represents a pixel within that patch. Here, \(I_k^{\text{render}}\) is the color of the rendered image at pixel \(k\), while \(I_k^{t_n}\) refers to the ground truth color at pixel \(k\) from the target timestamp \(t_n\). \(I_p^{\text{render}}\) and \(I_p^{t_n}\) are collections of pixels within patch \(p\).

\subsubsection{Iterative Confidence Expansion.}
To refine confidence maps and expand reliable regions, we exploit the observation that 3DGS typically renders higher-quality images near the training viewpoints. For each viewpoint \(i\) in \(t_0\), we render \(\mathbf{I}_{\text{render}}^i\) using the current model \(G_n\) and compute a fine-grained similarity score against the ground-truth image \(\mathbf{I}_{t_0}^i\).

We refine the confidence map through finer-grained similarity evaluation and stricter thresholds. Patch-wise modified SSIM scores are used to progressively expand confident regions across optimization iterations. We define the updated confidence map \(C_{\text{iter}}^i\) for viewpoint \(i\) as:
\begin{equation}
C_{\text{iter}}^i(p) = \begin{cases} 
1 & \text{if \;} mSSIM(p) \geq \tau_{\text{iter}} \\ 
0 & \text{otherwise}.
\end{cases}
\end{equation}

This finer partitioning facilitates localized expansion of high-confidence regions. The scene areas corresponding to these patches are added to the supervision set, and \(G_n\) is re-optimized with an updated loss.

As more stable regions are incorporated, the refined geometry of \(G_n\) improves rendering quality for viewpoints near \(t_0\), thus reinforcing a positive feedback loop. 
The refinement process terminates either after a fixed number of iterations or once the confidence map converges. 

At last, static regions are fixed based on high-confidence priors from $t_0$, while dynamic regions are optimized using sparse views from $t_n$.
The complete pipeline achieves geometric consistency in static regions while recovering fine details in dynamic areas under sparse conditions.

This progressive approach effectively bridges the sparse observations at \(t_n\) with the rich priors from \(t_0\), enabling efficient and accurate scene updates for real-world applications. The iterative confidence expansion ensures that the system adapts automatically to scene changes without manual intervention, striking a balance between reconstruction quality and computational efficiency.

\section{Experiments}
\textbf{Compared Methods:}
We compare our method with three baselines: original 3DGS method, a 3DGS-based sparse-view reconstruction approach InstantSplat \cite{fan2024instantsplat} and a recent scene editing framework.

The baseline is defined as direct 3DGS \cite{kerbl20233d} trained from scratch using only the sparse-view images at time $t_n$, without leveraging image prior information from time $t_0$, yet utilise the fused dense point clouds in our method. This setup reflects a typical reconstruction pipeline under sparse and temporally isolated observations, and serves as a strong lower bound for performance in the absence of scene priors.
The GaussianEditor \cite{chen2024gaussianeditor} comparison involves using a 3D scene editing method that enables semantic text-guided updates over an existing 3DGS scene. While it is not originally intended for cross-temporal reconstruction, we adapt it for this task by supplying annotated text prompts to indicate desired scene changes. This comparison highlights the limitations of editing-based approaches in achieving geometrically consistent and temporally accurate updates.

\textbf{Implement Details:}
All experiments are conducted on a single NVIDIA RTX A5000 GPU. We adopt the official 3DGS implementation as the reconstruction backbone and apply our method to both synthetic and real-world datasets.

For \textbf{Interference-Based Confidence Initialization}, each image is divided into $16\times16$ patches, and a patch is assigned high confidence if the modified SSIM score exceeds a threshold $\tau = 0.8$. Confidence maps are computed after a brief adaptation phase of 500 optimization steps using sparse-view images from the target timestamp $t_n$.

During \textbf{Progressive Cross-Temporal Optimization}, confidence maps are refined every 108 epochs using a stricter threshold $\tau_{\text{iter}} = 0.92$ and finer patch size $32\times32$. The full optimization process runs up to 7000 iterations or terminates earlier if the confidence map coverage change falls below 2\%. 

Each scene includes a pretrained 3DGS model at time $t_0$ (trained for 7000 steps). The update process at time $t_n$ uses 8 sparse training views and 4 test views. 

\textbf{Datasets:} 
We conducted experiments on our new dataset, which is specifically made for the cross-temporal scene update task. The dataset is well-suited for that task because it handles object additions, removals, and modifications within the scene, ensuring cross-temporal updates for the scene. 
The dataset consists of one virtual scene and four real-world scenes, namely City, Airpods, Cup, Keyboard, and Tower. At time $t_0$, each scene includes 100 images with corresponding camera poses, while at time $t_n$, it contains 12 images. All images are at standard 1920*1080p.
Additionally, the dataset includes a pre-trained 3DGS model with 7000 epochs trained at $t_0$, providing a structured benchmark for evaluating cross-temporal scene updates.

\textbf{Metrics:} We report image metrics of PSNR \cite{5596999}, SSIM \cite{wang2004ssim}, and LPIPS \cite{zhang2018lpips}, that cover different aspects of image quality for evaluation. We also report the time consumption for the training process.
PSNR (Peak Signal-To-Noise Ratio) compares a given signal with the source signal. LPIPS (Learned Perceptual Image Patch Similarity) aims to simulate perceptual similarity in human vision. SSIM (Structure Similarity Index Measure) is used to quantify structural similarity.

\begin{table}[t]
\centering
\label{tab:quantity}
\begin{tabular}{lcccc}
\toprule
\textbf{Method} & \textbf{PSNR}$\uparrow$ & \textbf{SSIM}$\uparrow$ & \textbf{LPIPS}$\downarrow$ & \textbf{Time/s}$\downarrow$ \\
\midrule
Baseline              & 15.87 & 0.683 & 0.447 & \textbf{160.55} \\
Instant-Splat         & 16.18 & 0.580 & 0.534 & 405.66 \\
GS-Editor             & 12.16 & 0.554 & 0.714 & 501.60 \\
Ours         & \textbf{23.89} & \textbf{0.8641} & \textbf{0.2394} & 310.00 \\
\bottomrule
\end{tabular}
\caption{Quantitative comparisons on the cross-temporal dataset.}
\label{tab:quantity}
\end{table}

\begin{figure*}[t]
\centering
\includegraphics[width=0.95\linewidth]{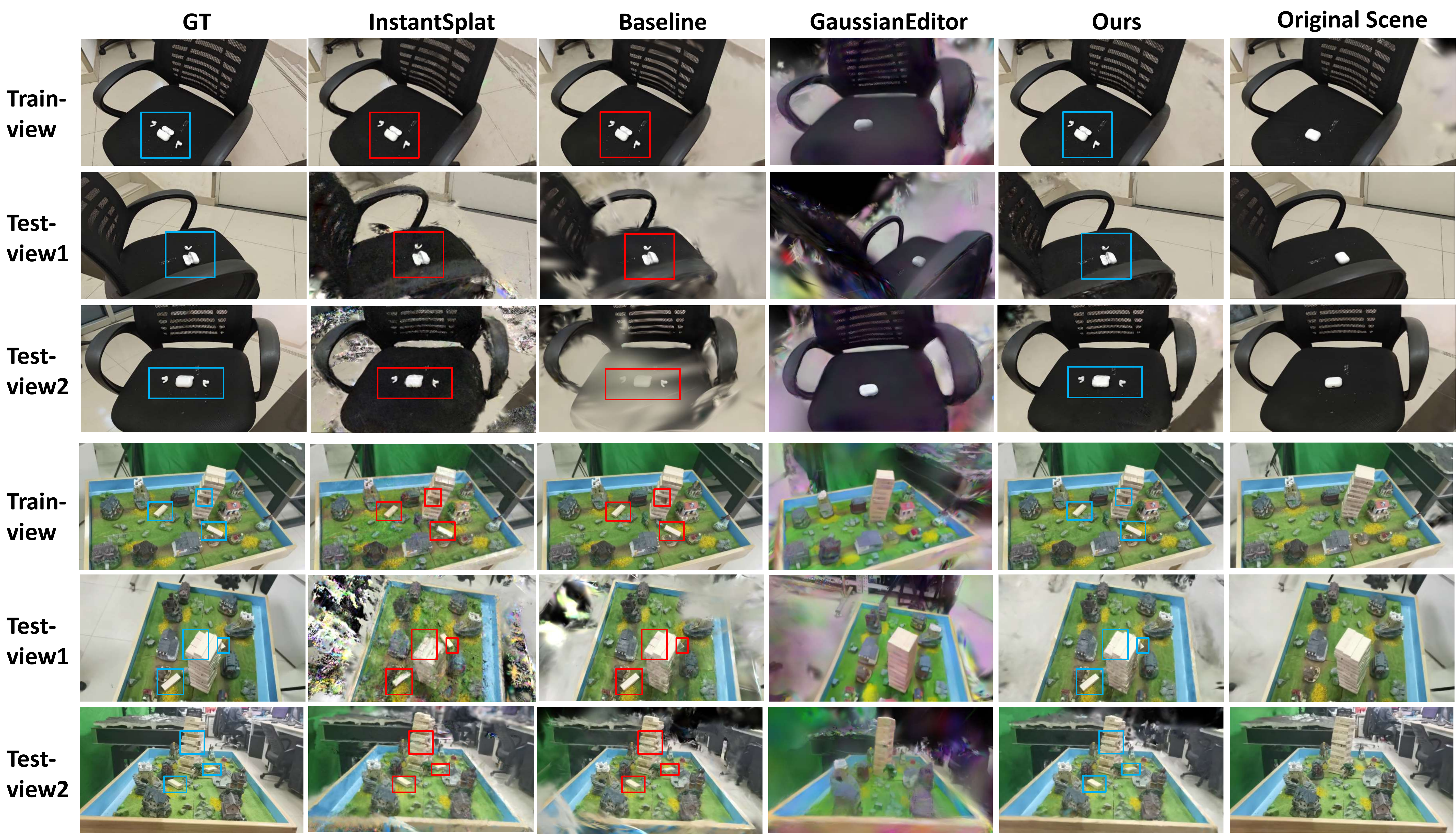}
\caption{Qualitative Comparison. We present the performance of four different methods (InstantSplat, Baseline, GaussianEditor, Ours) across three scenes in the dataset (Airpods, Tower). For each scene, we show one training view and two testing views to illustrate the effectiveness of each approach. The changed parts were highlighted in boxes.}
\label{fig:qulity}
\end{figure*}

\subsection{Quantitative Comparisons}

We compare our method with three baselines: (1) a standard 3DGS model trained from scratch using only sparse views at time $t_n$, (2) GaussianEditor, a 3DGS-based semantic editing framework, and (3) InstantSplat, a recent sparse-view reconstruction method.

Table~\ref{tab:quantity} presents the average performance across all scenes. Our method significantly outperforms all baselines on three quality metrics: PSNR, SSIM, and LPIPS. In particular, our approach achieves a PSNR improvement of over 7 dB compared to the standard 3DGS baseline, along with notable gains in perceptual similarity (LPIPS) and structural consistency (SSIM). These results demonstrate the effectiveness of leveraging historical priors and progressive optimization for cross-temporal scene updates.

While InstantSplat achieves faster rendering and acceptable quality under extreme sparsity, its performance is notably lower in fidelity-oriented metrics, indicating a trade-off between speed and reconstruction accuracy. GaussianEditor, originally designed for semantic-guided editing, struggles to produce geometrically consistent updates from sparse-view observations and suffers from degraded perceptual quality.

In terms of runtime, our method maintains a favorable balance between reconstruction quality and computational cost. Although slightly slower than the baseline, it is significantly more efficient than GaussianEditor and comparable to InstantSplat, with all methods completing within practical time bounds. Overall, our framework provides a scalable and accurate solution for cross-temporal 3D scene updates from sparse views.

\subsection{Qualitative Comparisons}

Figure~\ref{fig:qulity} illustrates qualitative results comparing four methods: our Cross-Temporal 3DGS, the direct 3DGS baseline, InstantSplat, and GaussianEditor. Each row shows a training view and two testing views, allowing visual inspection of reconstruction consistency, structural fidelity, and ability to reflect scene changes.

The baseline 3DGS model, trained from scratch using sparse views, often exhibits floating artifacts, structural collapse, or over-smoothed regions due to the lack of strong supervision. InstantSplat achieves fast reconstruction but suffers from coarse geometry and incomplete details, especially in occluded or updated areas.

GaussianEditor performs semantic-guided editing over a pre-trained scene, but the changes are often inconsistent across views and lack geometric precision, as it does not explicitly align historical and current observations.

In contrast, our method achieves superior structural completeness and temporal coherence. By leveraging high-confidence priors and progressive optimization, our reconstructions preserve fine details in static regions while faithfully integrating changes in dynamic parts. The consistent quality across testing views highlights the benefit of cross-temporal alignment and confidence-aware supervision.
These results confirm that our approach not only improves perceptual quality but also ensures view-consistent and structurally plausible updates under sparse observations.

\subsection{Ablation Study}

To demonstrate the effectiveness of each part of our proposed method, we conduct various ablation studies.

\begin{table}[t]
\centering
\label{tab:alignment_ablation}
\begin{tabular}{lccc}
\toprule
\textbf{Method} & \textbf{PSNR}$\uparrow$ & \textbf{SSIM}$\uparrow$ & \textbf{LPIPS}$\downarrow$ \\
\midrule
Ablation 1  & 23.67 & 0.860 & 0.258 \\
Ablation 2  & 21.84 & 0.808 & 0.324 \\
Ours        & \textbf{23.89} & \textbf{0.864} & \textbf{0.239} \\
\bottomrule
\end{tabular}
\caption{Ablation of Cross-Temporal Camera Alignment.}
\label{tab:alignment_ablation}
\end{table}

\begin{table}[h]
\centering

\label{tab:confidence_ablation}
\begin{tabular}{lcccc}
\toprule
\textbf{Method} & \textbf{PSNR}$\uparrow$ & \textbf{SSIM}$\uparrow$ & \textbf{LPIPS}$\downarrow$ & \textbf{Time/s}$\downarrow$ \\
\midrule
Ablation 3              & 19.38 & 0.759 & 0.342 & 370.18 \\
Ablation 4              & 21.79 & 0.814 & 0.291 & \textbf{271.25} \\
Ours        & \textbf{23.89} & \textbf{0.864} & \textbf{0.239} & 310.00 \\
\bottomrule
\end{tabular}
\caption{Ablation of Confidence Initialization and Progressive Optimization.}
\label{tab:confidence_ablation}
\end{table}

\textbf{Ablation of Cross-Temporal Camera Alignment:} To evaluate the impact of our camera alignment strategy, we perform two variants of ablation. In \textbf{Ablation 1}, we directly utilize the camera poses obtained from SfM at time $t_n$ without any temporal alignment. This results in misaligned coordinate systems between $t_0$ and $t_n$, introducing geometric inconsistencies that degrade the effectiveness of prior transfer and subsequent reconstruction. In \textbf{Ablation 2}, we partially apply our alignment procedure by estimating a coarse transformation but skip the ICP refinement step. This fails to correct fine misalignments, especially around detailed structures. As shown in Table~\ref{tab:alignment_ablation}, both variants lead to performance drops compared to the full method. Ablation 1 achieves a PSNR of 23.67, while Ablation 2 drops further to 21.84, along with a noticeable decrease in SSIM and increase in LPIPS. These results indicate that full alignment, including both coarse registration and ICP refinement, is critical for achieving accurate geometric consistency across time. Visual inspections further confirm that unaligned models often suffer from distorted structures and unstable projections.

\textbf{Ablation of Confidence Initialization and Progressive Optimization:} We also evaluate the contributions of our confidence-aware update strategy by conducting ablations on the two key components: confidence initialization and progressive confidence refinement. In \textbf{Ablation 3}, we disable the confidence estimation entirely and directly fine-tune the pretrained model $G_0$ on the sparse images at $t_n$. Direct finetuning of $G_0$ on $t_n$ images without confidence guidance yields significantly lower image quality and unstable training, confirming the effectiveness of our confidence-aware approach. In \textbf{Ablation 4}, we retain the initial confidence maps but skip progressive updates, keeping them fixed throughout training. This restricts the system’s ability to adaptively expand reliable supervision. As presented in Table~\ref{tab:confidence_ablation}, the full method consistently outperforms both ablations across all metrics. Compared to Ablation 3, our method improves PSNR by 4.5 dB and achieves better structural similarity and perceptual quality. Compared to Ablation 4, the gains are more modest but consistent, particularly in LPIPS (0.2394 vs. 0.291), indicating better perceptual alignment. Furthermore, the computational overhead introduced by confidence refinement is minimal (approx.~30 seconds), due to our confidence update strategy. These results confirm that both initial confidence estimation and its progressive refinement are essential for stable and effective cross-temporal optimization.

\textbf{Ablation of Sparse Camera View Quantity and Distribution:} To analyze how the number and spatial distribution of sparse views affect reconstruction performance, we perform a systematic ablation by varying the number of input views ($4$, $6$, $8$) and comparing two spatial layouts: \textit{concentrated} (views clustered in one direction) and \textit{uniform} (evenly distributed views). This simulates real-world scenarios where images may be limited in quantity and biased in viewpoint. Table~\ref{tab:view_ablation} reports the quantitative results. Increasing the number of views improves reconstruction quality, with diminishing returns beyond 6 views. Under each view count, uniformly distributed views consistently outperform concentrated ones. The effect is most prominent in the LPIPS metric, where uniform layouts reduce perceptual distortion. Notably, even with only 4 uniform views, our method maintains competitive quality (PSNR 23.32, LPIPS 0.2629), demonstrating its robustness to extreme sparsity. These results emphasize the importance of both view quantity and spatial coverage when designing data collection protocols for cross-temporal updates.

\begin{table}[t]
\centering

\label{tab:view_ablation}
\begin{tabular}{c|c|ccc}
\toprule
\textbf{Views} & \textbf{Layout} & \textbf{PSNR}$\uparrow$ & \textbf{SSIM}$\uparrow$ & \textbf{LPIPS}$\downarrow$ \\
\midrule
\multirow{2}{*}{4} 
 & Concentrated  & 23.26 & 0.854 & 0.263 \\
 & Uniform  & 23.32 & 0.855 & 0.262 \\
\addlinespace
\multirow{2}{*}{6} 
 & Concentrated  & 23.55 & 0.853 & 0.263 \\
 & Uniform  & 23.56 & 0.860 & 0.251 \\
\addlinespace
\multirow{2}{*}{8} 
 & Concentrated & 23.79 & 0.859 & 0.263 \\
 & Uniform  & 23.89 & 0.864 & 0.239 \\

\bottomrule
\end{tabular}
\caption{Ablation of Sparse Camera View Quantity and Distribution.}
\label{tab:view_ablation}
\end{table}

\section{Limitation and Future Work}
Our method still exhibits several limitations. Under extreme sparsity, DUSt3R poses may become unreliable due to limited image overlap or weak textures, which can propagate errors to the initialization and subsequent optimization. Moreover, handling drastic geometric changes across time also remain challenging. Overall, our approach assumes predominantly stable global scene layouts; large-scale deformations and severe structural changes pose challenges and are intended for future work.

\section{Conclusion}
In this paper, we present \textbf{Cross-Temporal 3DGS}, a unified framework for updating or reconstructing 3D scenes across different timestamps using sparse-view images and historical priors. Our method tackles the core challenges of cross-temporal modeling through three components: cross-temporal camera alignment, interference-based confidence initialization, and progressive optimization for integrating priors over time. Experiments on real and synthetic datasets verify the contribution of each component and show that our approach achieves high-fidelity, temporally consistent reconstructions even under severe sparsity. By enabling reliable updates from as few as 4 unconstrained images, Cross-Temporal 3DGS offers a practical and data-efficient solution for long-term scene modeling.

\section{Acknowledgement}
This work was supported by the National Natural Science Foundation of China (Project Number: 62272019), the China Postdoctoral Science Foundation under Grant Number 2025M774236, the Postdoctoral Fellowship Program of CPSF under Grant Number GZC20242159 and ``the fundamental research funds for the central universities".

\bibliography{aaai2026}

\end{document}